\documentclass[sigconf]{acmart}

\AtBeginDocument{
  }

\copyrightyear{2026}
\acmYear{2026}
\setcopyright{cc}
\setcctype{by}
\acmConference[CIKM '26]{Proceedings of the 35th ACM International Conference on Information and Knowledge Management}{November 07--11, 2026}{Rome, Italy}
\acmBooktitle{Proceedings of the 35th ACM International Conference on Information and Knowledge Management (CIKM '26), November 07--11, 2026, Rome, Italy}
\acmDOI{10.1145/3799682.3840891}
\acmISBN{979-8-4007-2539-5/2026/11}
\usepackage{algorithm}
\usepackage{algorithmic}
\usepackage{multirow}
\usepackage{colortbl}
\usepackage{enumitem}

\begin{document}

\title{MCTS-KBQA: Monte Carlo Tree Search with Information Gain Rewards for Knowledge Base Question Answering}

\author{Guanming Xiong}
\email{gm\string_xiong@pku.edu.cn}
\affiliation{
  \institution{Peking University}
  \city{Beijing}
  \country{China}}

\author{Haochen Li}
\email{haochenli@pku.edu.cn}
\affiliation{
  \institution{AlignBase}
  \city{Beijing}
  \country{China}}

\author{Zonghong Dai}
\authornote{Corresponding author.}
\email{zhdai23@m.fudan.edu.cn}
\affiliation{
  \institution{Fudan University}
  \city{Shanghai}
  \country{China}}

\author{Liqiang Wen}
\email{wenlq@pku.edu.cn}
\affiliation{
  \institution{Peking University}
  \city{Beijing}
  \country{China}}

\author{Wen Zhao}
\email{zhaowen@pku.edu.cn}
\affiliation{
  \institution{Peking University}
  \city{Beijing}
  \country{China}}

\renewcommand{\shortauthors}{Guanming Xiong, Haochen Li, Zonghong Dai, Liqiang Wen, and Wen Zhao}

% --------------------------------------------- #

\begin{abstract}
This work investigates how to improve large language model (LLM)-based reasoning for knowledge base question answering (KBQA) via Monte Carlo Tree Search (MCTS). Applying MCTS to LLM-based KBQA remains challenging because reward design is difficult and rollout-based search is computationally expensive. Existing MCTS-style methods either rely on direct LLM scoring or require substantial data to train separate reward models, and they often provide rewards only at terminal states. To address these limitations, we propose Fast MCTS, which replaces terminal rollouts with an information gain (IG) reward for intermediate states. The IG reward is implemented as a question-conditioned PPL-ratio proxy over sanitized interaction histories, computed by forward passes of an open-source instruction LLM without additional reward-model training. Experiments on four KBQA benchmarks show that Fast MCTS consistently outperforms linear baselines and generally improves the accuracy-cost trade-off relative to rollout-based Classic MCTS.\footnote{Code and data are available at \url{https://github.com/JimXiongGM/MCTS-KBQA}.}
\end{abstract}

    


\maketitle

\section{Introduction}

Knowledge base question answering (KBQA) aims to answer natural language (NL) questions over structured knowledge bases (KBs).
Powered by large language models (LLMs) such as ChatGPT \cite{Ouyang-Long-NeurIPS-2022-InstructGPT} and GPT-4 \cite{OpenAI-2023-GPT4}, recent KBQA methods often view LLMs as agents and KBs as environments \cite{Gu-Yu-EMNLP-2024-Middleware,Jiang-Jinhao-arXiv-2024-KG-Agent,Xiong-Guanming-ACL-2024-Interactive-KBQA}.

Monte Carlo Tree Search (MCTS) has recently shown strong performance in reasoning tasks such as mathematical problem solving and code generation.
By iterating selection, expansion, simulation, evaluation, and backpropagation, MCTS can explore multiple reasoning branches instead of committing to a single linear trajectory.
However, applying it to LLM-based KBQA faces two challenges: reward design and rollout cost.
Some studies \cite{Chen-Guoxin-NIPS-2024-AlphaMath,Zhang-Di-arXiv-2024-LLaMA-Berry,Zhou-Andy-ICML-2024-Language-Agent-Tree-Search} obtain rewards directly from LLMs via prompting, while others \cite{Zhang-Dan-NIPS-2024-ReST-MCTS,Lightman-Hunter-arXiv-2023-ProcessSupervision,Luo-Haoran-arXiv-2025-KBQA-o1} train dedicated reward models.
However, prompt-based rewards can be difficult to calibrate \cite{Huang-Jie-ICLR-2024-LLM-Cannot-Self-Correct,Jiang-Dongwei-AAAI-2025-SELF-INCORRECT}, and trained reward models require extra supervision.
Classic MCTS is also inefficient because each rollout evaluates only a terminal state while discarding intermediate steps that may already reveal useful KB evidence.

To explicitly value intermediate progress on topic-entity localization, predicate acquisition, and constraint construction, we propose an information gain (IG) reward for MCTS.
KBQA tool observations often expose partial semantic evidence before a full SPARQL query is completed, making intermediate states informative.
Our reward requires no fine-tuning, is computed by forward passes of open-source LLMs, and provides immediate feedback, allowing Fast MCTS to avoid terminal rollouts.

Our contributions are as follows:
\begin{itemize}[nolistsep]
    \item We introduce MCTS as a drop-in replacement for the linear inference algorithm in the SFT agent of Interactive-KBQA \cite{Xiong-Guanming-ACL-2024-Interactive-KBQA}, yielding significant gains under the same low-resource settings.
    \item We propose an IG reward that is computed by forward passes of open-source LLMs without additional reward-model training, while remaining effective as a process-level branch-ranking proxy.
    \item We develop a Fast MCTS variant that provides immediate feedback at intermediate steps, improving the accuracy-cost trade-off relative to rollout-based Classic MCTS on most datasets while remaining comparable on heterogeneous KQA Pro.
\end{itemize}

\begin{figure*}[t]
    \centering
    \includegraphics[width=1\textwidth,page=1]{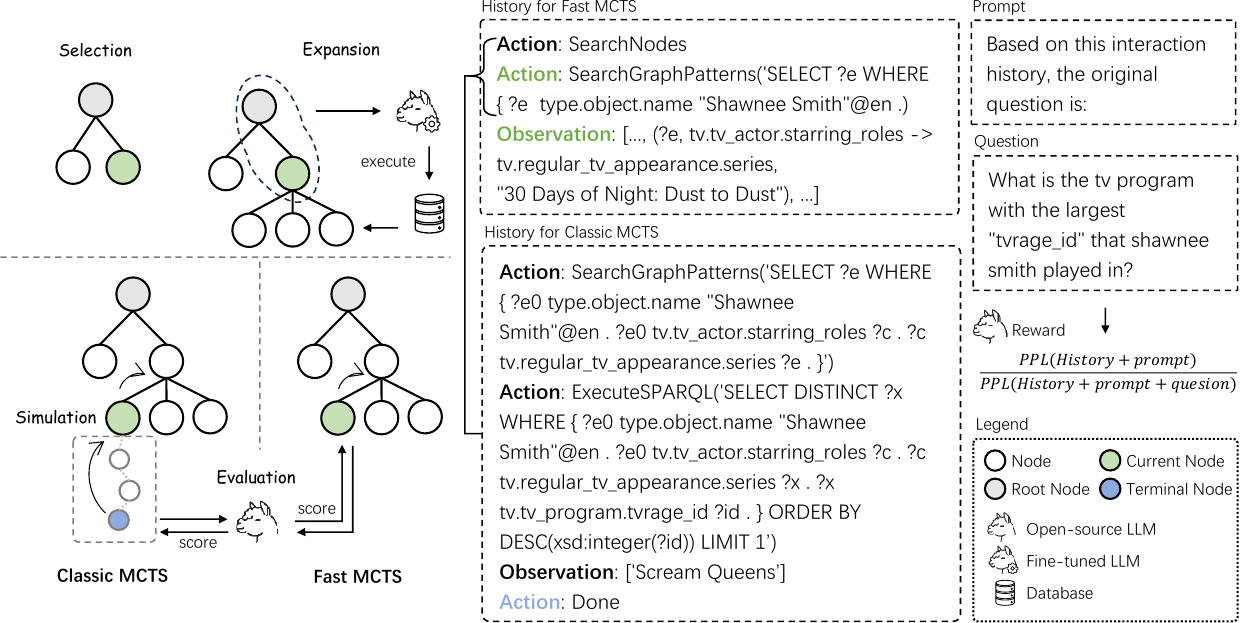}
    \caption{The left panel illustrates the distinction between Classic MCTS and Fast MCTS, while the right panel provides a concrete example of reward calculation.}
    \label{fig:overview}
\end{figure*}

\section{Related Work}

Recent KBQA methods largely follow the \textbf{agent-environment framework}, conceptualizing LLMs as agents and KBs as environments.
\citet{Gu-Yu-ACL-2023-Pangu} emphasized plan-plausibility evaluation in few-shot settings, while \citet{Li-Tianle-ACL-2023-KB-BINDER} generated logical forms before refining them into executable queries.
\citet{Jiang-EMNLP-2023-StructGPT} developed specialized KB interfaces, while \citet{Gu-Yu-EMNLP-2024-Middleware} and \citet{Liu-Xiao-ICLR-2023-AgentBench} designed tool sets for agent-environment interaction.
\citet{Jiashuo-Sun-ICLR-2024-Think-on-Graph} introduced iterative beam-search reasoning on KBs.
These methods improve tool use and local planning, but explicit tree search remains relatively underexplored in low-resource KBQA.

\textbf{MCTS-based reasoning} has shown strong results on tasks such as mathematical reasoning. Its key difficulty is reward design.
One line of work derives rewards directly from LLMs.
\citet{Hao-Shibo-EMNLP-2023-Reasoning-via-Planning-RAP} used LLMs as world models to simulate future states, and \citet{Zhang-Di-arXiv-2024-Accessing-GPT-4-level-Mathematical} combined self-refinement with direct reward prompting.
However, \citet{Zhang-Di-arXiv-2024-LLaMA-Berry} reported substantial variance in such generative rewards, while \citet{Huang-Jie-ICLR-2024-LLM-Cannot-Self-Correct,Jiang-Dongwei-AAAI-2025-SELF-INCORRECT} found that LLMs cannot reliably self-evaluate or self-correct.
Another line trains reward models.
Given question-answer pairs, these methods use MCTS to discover reasoning paths and score intermediate steps under weak supervision.
\citet{Zhang-Dan-NIPS-2024-ReST-MCTS} proposed a distance-based reward rule, while \citet{Chen-Guoxin-NIPS-2024-AlphaMath} added linear layers to jointly train the value model and reasoning LLM.
\citet{Xie-Yuxi-arXiv-2024-MCTS-Preference-Learning} further used step-wise preference data with Direct Preference Optimization (DPO), and \citet{Wang-Chaojie-arXiv-2024-Q-star} proposed step-wise value preference optimization.
However, these approaches require substantial annotation, and their rule-based intermediate supervision is often coarse.

Closer to our setting, \citet{Zhou-Andy-ICML-2024-Language-Agent-Tree-Search} treated LLMs as agents and used self-generated and self-consistency scores as rewards, but assumed ground-truth feedback from the environment, which is unrealistic in practice.
\citet{Luo-Haoran-arXiv-2025-KBQA-o1} also applied MCTS to KBQA, but their method jointly trains policy and reward models and evaluates only final states, providing no feedback on intermediate steps.

Therefore, designing effective reward functions and MCTS frameworks for KBQA remains an open challenge.

\section{Approach}

\subsection{Task Formulation and Search State}

We adapt MCTS to KBQA and introduce an information gain (IG) reward as an efficient, interpretable value proxy for intermediate states.
We further propose Fast MCTS to reduce inference cost. Figure \ref{fig:overview} summarizes the framework and the reward computation.

\paragraph{KBQA Formulation.}
A knowledge base (KB) is modeled as a set of triples $\mathcal{K} \subseteq \mathcal{E} \times \mathcal{P} \times (\mathcal{E} \cup \mathcal{C} \cup \mathcal{L})$, where $\mathcal{E}$ denotes the set of entities, $\mathcal{P}$ represents the set of predicates (including relations and qualifiers), $\mathcal{C}$ denotes the set of classes (e.g., concepts), and $\mathcal{L}$ comprises the set of literal values.
Given a question $Q$ and a knowledge base $\mathcal{K}$, our goal is to generate an executable SPARQL expression that answers the question.
The semantic parsing process can be formalized as $p(\text{SPARQL} | Q, \mathcal{K})$.

\paragraph{Interaction State.}
A node $e=(t, a, o)$ contains a thought process $t$, an action $a$, and an observation $o$, except for the root node $e_{root}=Q$ which contains only the question.
The state $s$ of a node $e$ is defined as the path from the root node to $e$, formally represented as:

\begin{equation}
\begin{aligned}
    s_e &= (e_{root}, e_1, ..., e_d) \\
        &= (Q, t_1, a_1, o_1, ..., t_d, a_d, o_d)
\end{aligned}
\end{equation}

where $d$ is the depth of node $e$ in the tree, and the second line flattens each $e_i=(t_i,a_i,o_i)$ into its three components. Thus, the state of a node represents the complete history of interactions along the branch.

\paragraph{Action Space.}
Following Interactive-KBQA, each action is selected from \{SearchNodes, SearchGraphPatterns, ExecuteSPARQL, Done\}. SearchNodes locates topic entities, SearchGraphPatterns retrieves schema patterns and predicates, ExecuteSPARQL executes a candidate query, and Done terminates a trajectory. Executing an action produces an observation, which is attached to the corresponding child node.
For GrailQA, we additionally include SearchTypes to retrieve candidate types required by its schema-level generalization setting.

\subsection{Classic MCTS for KBQA}

We first describe a classic MCTS pipeline for KBQA to separate generic tree-search mechanics from our contributions.
Classic MCTS iteratively performs Selection, Expansion, Simulation (rollout), Evaluation, and Backpropagation until termination criteria are met.

\paragraph{Selection.}
Starting from the root, we select a leaf using the UCT rule \cite{Kocsis-ECML-2006-UCT} balancing exploration and exploitation:
\begin{equation}
    \text{UCT}(e) = \frac{w_e}{n(e)} + \sqrt{2\,\frac{\log N(e)}{n(e)}}.
\end{equation}

Here $w_e$ is accumulated reward, $n(e)$ the visit count of node $e$, and $N(e)$ that of its parent. We pick $e_{select} = \arg\max_{e \in E} \text{UCT}(e)$ over non-terminal nodes $E$.

\paragraph{Expansion.}
Given $e_{select}$ we query the LLM agent to sample $n$ candidate actions:
\begin{equation}
\{a_i\}_{i=1}^n = \operatorname{Actor}(s_{e_{select}}),
\end{equation}
where each $a_i$ is a triple of a thought, an action, and its parameters. Executing $a_i$ yields an observation $o_i$, forming a child node $c_i=(t_i,a_i,o_i)$ consistent with the node definition above.

\paragraph{Simulation (Rollout).}
In the classic setting, for each newly created non-terminal child we run a stochastic rollout that continues along a single path until a terminal state (Done or maximum depth) is reached; intermediate states visited during the rollout are transient and are not added to the persistent tree. The only evaluative signal produced by a rollout is the terminal reward at its end, which is then backpropagated to update the nodes on the selected path.

\paragraph{Evaluation.}
A generic reward function assigns a scalar to the terminal state returned by simulation. For a terminal child, $\text{EvalTerminal}(\cdot)$ applies this reward function directly to the terminal state without requiring gold answers at inference time; the resulting value is then propagated upward.

\paragraph{Backpropagation.}
Rewards are accumulated with a clipped depth-aware linear decay to discourage overly deep, low-yield branches while still propagating informative signals upward. Let $d_e$ be the depth of node $e$, $d_{exp}$ an expected (preferred) depth and $r_e$ the immediate reward (from terminal simulation in the classic setting). We update
\begin{equation}
\label{eq:backprop}
    w_e \leftarrow w_e + r_e \cdot \max\!\left(0, 1 - \gamma \cdot \max(0, d_e - d_{exp})\right),
\end{equation}
where $\gamma$ is a decay coefficient. This linear schedule preserves comparability across sibling branches while softly penalizing over-deep trajectories, and the outer clipping prevents negative reward propagation for very deep states.

\paragraph{Termination.}
Search halts when (i) a simulation budget is exhausted, (ii) maximum depth is reached, or (iii) a target number $k$ of valid terminal branches (producing executable, non-empty SPARQL answers) is collected. Final answers are chosen by majority vote over execution results, reducing variance from any single trajectory.

\subsection{Information Gain Reward}

\paragraph{Motivation.}
Classic MCTS is inefficient for KBQA because rewards are available only at terminal states. We instead score intermediate states by measuring how much the interaction history helps predict the question, using only forward passes.
Entity localization, predicate acquisition, and SPARQL construction reveal complementary aspects of the question, allowing the score to distinguish promising partial states before a full trajectory is completed.

\paragraph{Sanitized History.}
Let $Q$ denote the question, $H = (t_1,a_1,o_1,\ldots,t_d,a_d,o_d)$ the raw history, and $H'=(a'_1,\ldots,a'_d,o_d)$ the \emph{sanitized} history. Here, free-form thoughts and lexical parameters that may directly copy $Q$ are removed, while the executed actions and the current observation are preserved. Unsanitized histories are not valid reward inputs in our setting because copied question spans or free-form paraphrases would leak the target question into the score. Perplexity (PPL) measures how predictable a sequence is under a frozen language model; lower PPL indicates higher probability. We write $\operatorname{PPL}(H',Q)$ for the perplexity of the concatenated sequence obtained by appending $Q$ after $H'$ (a formal definition is given in Appendix~\ref{app_sec:ig_reward_details}).

\paragraph{MI-inspired Proxy.}
Motivated by the mutual information (MI) between $Q$ and $H'$, we define the reward as
\begin{equation}
    \label{eq:reward_mapping}
    r(H', Q) = \alpha\, \sigma\!\left(\beta \log \frac{\operatorname{PPL}(H')}{\operatorname{PPL}(H',Q)}\right),
\end{equation}
where $\sigma(x)=\frac{1}{1+e^{-x}}$, and $\alpha,\beta$ calibrate the reward scale. For brevity, we refer to this score as the \emph{IG reward} throughout the paper. It is a question-conditioned PPL-ratio surrogate for local branch ranking, motivated by MI, rather than an exact mutual-information estimate; the length-aware motivation is provided in Appendix~\ref{app_sec:ig_reward_details}.

\paragraph{Reward Scope.}
Sanitization prevents trivial lexical overlap from inflating the score, so the reward reflects reasoning progress rather than string matching. We use the IG score as a sibling-level branch-ranking proxy rather than a calibrated estimator of terminal reward. Section~\ref{sec:ig_validation} shows that the resulting ranking preserves promising branches well enough for Fast MCTS to replace expensive simulations with two forward passes.

\subsection{Fast MCTS for KBQA}
\label{sec:fast_mcts}

\begin{algorithm}[t]
    \small
    \begin{algorithmic}[1]
 \STATE Initialize root with question $Q$
 \WHILE{termination criteria not met}
    \STATE $e \gets \text{UCT}(\text{root})$ \COMMENT{Selection}
    \STATE Sample $\{a_i\}_{i=1}^n \gets \text{Actor}(s_e)$ \COMMENT{Expansion}
    \FOR{each $a_i$}
  \STATE $o_i \gets \text{ToolExecute}(a_i)$ \COMMENT{KB interaction}
  \STATE Create child $c_i=(t_i,a_i,o_i)$
  \IF{$c_i$ non-terminal}
    \STATE $R_i \gets \text{Simulate}(c_i)$ \COMMENT{Rollout to terminal}
  \ELSE
    \STATE $R_i \gets \text{EvalTerminal}(c_i)$ \COMMENT{Apply reward}
  \ENDIF
  \STATE Backpropagate($c_i, R_i$) \COMMENT{Eq.~\ref{eq:backprop}}
     \ENDFOR
 \ENDWHILE
 \RETURN MajorityVote(valid terminal answers)
    \end{algorithmic}
    \caption{Classic MCTS for KBQA}
    \label{alg:classic_mcts}
\end{algorithm}

\begin{algorithm}[t]
    \small
    \begin{algorithmic}[1]
 \STATE Initialize root with $Q$
 \WHILE{termination criteria not met}
     \STATE $e \gets \text{UCT}(\text{root})$
     \STATE Sample $\{a_i\}_{i=1}^n \gets \text{Actor}(s_e)$
     \FOR{each $a_i$}
  \STATE $o_i \gets \text{ToolExecute}(a_i)$
  \STATE $c_i=(t_i,a_i,o_i)$
  \STATE $H'_i \gets \text{Sanitize}(s_{c_i})$ \COMMENT{Remove thoughts \& copy-prone params}
  \STATE $r_i \gets \text{IGReward}(H'_i)$ \COMMENT{Two LM passes}
  \STATE Backpropagate($c_i, r_i$)
     \ENDFOR
 \ENDWHILE
 \RETURN MajorityVote(valid terminal answers)
    \end{algorithmic}
    \caption{Fast MCTS for KBQA}
    \label{alg:fast_mcts}
\end{algorithm}

Algorithms~\ref{alg:classic_mcts} and \ref{alg:fast_mcts} summarize Classic MCTS and Fast MCTS.
The key change is to replace simulation-based evaluation with direct IG rewards. In Classic MCTS, each newly expanded non-terminal child triggers a rollout to obtain a terminal reward. In Fast MCTS, we instead compute $r(H', Q)$ from Eq.~\ref{eq:reward_mapping} using two forward passes on $H'$ and $(H',Q)$, providing immediate feedback without terminal simulation.
The same IG reward is used for all expanded children, including terminal children; terminal states are still retained as valid candidates for the final majority vote when they produce executable non-empty answers.

\paragraph{Computational Cost.}
Classic MCTS incurs high cost per expansion: each rollout requires multiple autoregressive Actor calls until reaching a terminal state. Let $C_{\text{gen}}(L)$ denote the cost of generating $L$ tokens and $C_{\text{fwd}}(L)$ the cost of a forward pass over $L$ tokens. A rollout incurs cost $\sum_{i=1}^{d_{\text{roll}}} C_{\text{gen}}(L_i)$ where $d_{\text{roll}}$ is the rollout depth. In contrast, Fast MCTS computes IG rewards via two forward passes: $2C_{\text{fwd}}(|H'|)$. Therefore, Fast MCTS reduces rollout-generation cost relative to Classic MCTS, although its forward-pass reward evaluation can still make it slower than simple majority-vote linear inference in wall-clock time.

\section{Experiment}

\begin{table}[t]
\centering
\scalebox{1}{
\begin{tabular}{lcc}
    \toprule
    \textbf{Dataset} & \textbf{\begin{tabular}[c]{@{}c@{}}\#Instance\\(Ann/Test)\end{tabular}} & \textbf{\begin{tabular}[c]{@{}c@{}}Raw \#Instance\\(Train/Dev/Test)\end{tabular}} \\ \midrule
    WebQSP & 100 / 1,621 & 3,098 / - / 1,639 \\
    CWQ & 200 / 500 & 27,639 / 3,519 / 3,531 \\
    GrailQA & 150 / 500 & 44,337 / 6,763 / 13,231 \\
    KQA Pro & 450 / 900 & 94,376 / 11,797 / 11,797 \\ \bottomrule
\end{tabular}
}
\caption{Statistics of the datasets used in our experiments.}
\label{tab:data_statistics}
\end{table}

\begin{table*}[htp]
\centering
\scalebox{1}{
\begin{tabular}{lcccccccccc}
    \toprule
    \multicolumn{1}{c}{\multirow{2}{*}{\textbf{Method}}} & \multicolumn{4}{c}{\textbf{WebQSP}} & \multicolumn{6}{c}{\textbf{CWQ}} \\ \cline{2-11} 
    \multicolumn{1}{c}{} & \textbf{1-hop} & \textbf{2-hop} & \textbf{F1} & \textbf{RHits@1} & \textbf{Conj} & \textbf{Compo} & \textbf{Compa} & \textbf{Super} & \textbf{F1} & \textbf{EM} \\ \midrule
    % \textit{Reference Systems} &  &  &  &  &  &  &  &  &  &  \\
    DeCAF $\dagger$ & 76.48 & 74.80 & 75.87 & 81.21 & 63.74 & 39.82 & 31.16 & 21.21 & 49.12 & 57.80 \\
    \textit{Linear Decision} &  &  &  &  &  &  &  &  &  &  \\
    Prompting w/ GPT-4.1 & 70.50 & 69.16 & 70.02 & 69.67 & 51.08 & 38.27 & 63.20 & 50.00 & 46.15 & 55.40 \\
    SFT w/ open-LLM & 58.85 & 53.23 & 56.83 & 58.88 & 35.82 & 38.18 & 47.87 & 36.36 & 37.47 & 41.60 \\
    \quad - Majority Vote & 67.82 & 62.22 & 65.81 & 67.95 & 57.09 & 56.30 & 56.67 & 45.45 & 55.97 & 61.20 \\ \midrule
    \textit{Tree Search} &  &  &  &  &  &  &  &  &  &  \\
    BFS & 68.90 & 64.13 & 67.18 & 68.88 & 64.81 & 47.43 & 65.47 & 42.42 & 55.96 & 63.60 \\
    DFS & 63.84 & 59.14 & 62.15 & 63.80 & 62.37 & 50.04 & 47.14 & 44.70 & 55.19 & 64.20 \\
    Random & 69.46 & 62.66 & 67.01 & 68.50 & 62.33 & 48.50 & 68.67 & 43.94 & 55.54 & 66.00 \\
    IG-Beam Search & 67.04 & 64.68 & 66.19 & 68.29 & 69.49 & 54.66 & 58.13 & 42.42 & 60.82 & 68.20 \\
    Classic MCTS & 70.37 & 66.95 & 69.14 & 71.55 & 71.97 & 56.44 & \textbf{64.80} & 48.48 & 63.44 & 71.00 \\
    Fast MCTS & \textbf{75.67} & \textbf{70.66} & \textbf{73.87} & \textbf{75.43} & \textbf{80.41} & \textbf{62.12} & 64.56 & \textbf{55.05} & \textbf{70.15} & \textbf{80.20} \\
    \quad - w/ Const Reward & 69.24 & 60.82 & 66.21 & 67.78 & 65.77 & 52.13 & 51.20 & 46.97 & 57.99 & 67.40 \\ \midrule \midrule
    \multicolumn{1}{c}{\multirow{2}{*}{\textbf{Method}}} & \multicolumn{10}{c}{\textbf{KQA Pro}} \\ \cline{2-11} 
    \multicolumn{1}{c}{} & \textbf{Ct} & \textbf{QA} & \textbf{QAQ} & \textbf{QN} & \textbf{QR} & \textbf{QRQ} & \textbf{SA} & \textbf{SB} & \textbf{Vf} & \textbf{Acc} \\ \midrule
    % \textit{Reference Systems} &  &  &  &  &  &  &  &  &  &  \\
    BART-SPARQL $\dagger$ & 89.11 & 89.81 & 87.10 & 79.28 & 96.32 & 80.52 & 95.24 & 95.08 & 87.27 & 89.00 \\
    \textit{Linear Decision} &  &  &  &  &  &  &  &  &  &  \\
    Prompting w/ GPT-4.1 & 74.26 & 81.48 & 64.52 & 71.17 & 76.47 & 58.44 & 76.19 & 56.56 & 77.27 & 70.78 \\
    SFT w/ open-LLM & 67.33 & 79.63 & 78.49 & 62.16 & 90.44 & 58.44 & 71.43 & 85.25 & 76.36 & 75.78 \\
    \quad - Majority Vote & 77.23 & 82.41 & 83.87 & 78.38 & 92.65 & 83.12 & 80.95 & 91.80 & 75.45 & 83.44 \\ \midrule
    \textit{Tree Search} &  &  &  &  &  &  &  &  &  &  \\
    BFS & 71.29 & 82.41 & 79.57 & 81.08 & 90.44 & 72.73 & 80.95 & \textbf{97.54} & 76.36 & 82.33 \\
    DFS & 72.28 & 84.26 & 81.72 & 73.87 & 91.91 & 75.32 & 83.33 & 94.26 & 75.45 & 82.00 \\
    Random & 73.27 & 85.19 & 80.65 & 80.18 & 93.38 & 77.92 & 76.19 & 94.26 & 75.45 & 83.00 \\
    IG-Beam Search & 59.41 & 81.48 & 81.72 & 79.28 & 92.65 & 79.22 & 76.19 & 90.98 & 75.45 & 80.56 \\
    Classic MCTS & 75.25 & \textbf{88.89} & \textbf{87.10} & \textbf{85.59} & 92.65 & \textbf{83.12} & \textbf{85.71} & 94.26 & 82.73 & \textbf{86.67} \\
    Fast MCTS & \textbf{77.23} & 85.19 & 81.72 & 84.68 & \textbf{94.85} & 81.82 & 83.33 & 95.90 & \textbf{85.45} & 86.44 \\
    \quad - w/ Const Reward & 47.52 & 75.00 & 77.42 & 81.08 & 91.91 & 79.22 & 76.19 & 85.25 & 60.91 & 75.56 \\ \bottomrule
\end{tabular}
}
\caption{Main results on WebQSP, CWQ, and KQA Pro. $\dagger$: scores recomputed on our evaluation subsets using the authors' released predictions. ``w/ Const Reward'' is an ablation of Fast MCTS that replaces the IG score with a constant reward. Bold numbers mark the best low-resource result in each column.}
\label{tab:main_results}
\end{table*}

\begin{table}[htp]
\centering
\scalebox{1}{
\begin{tabular}{lcccc}
    \toprule
    \textbf{Method} & \textbf{i.i.d.} & \textbf{Comp.} & \textbf{Zero} & \textbf{F1} \\ \midrule
    % \textit{Reference Systems} &  &  &  &  \\
    DeCAF $\dagger$ & 69.42 & 51.46 & 61.31 & 61.17 \\
    KBQA-o1 & 85.50 & 77.60 & 76.10 & 78.50 \\
    \textit{Linear Decision} &  &  &  &  \\
    Prompting w/ GPT-4.1 & 62.05 & 58.24 & 67.22 & 63.92 \\
    SFT w/ open-LLM & 58.20 & 48.96 & 57.25 & 55.65 \\
    \quad - Majority Vote & 62.13 & 61.62 & 67.54 & 64.86 \\ \midrule
    \textit{Tree Search} &  &  &  &  \\
    BFS & 65.66 & 58.89 & 71.76 & 67.37 \\
    DFS & 59.02 & 60.75 & 73.18 & 66.85 \\
    Random & 66.91 & 63.81 & 75.56 & 70.77 \\
    IG-Beam Search & 78.47 & 73.34 & 82.36 & 79.38 \\
    Classic MCTS & 79.64 & 72.42 & \textbf{83.21} & 79.91 \\
    Fast MCTS & \textbf{80.30} & \textbf{74.61} & 83.13 & \textbf{80.52} \\
    \quad - w/ Const Reward & 66.97 & 61.20 & 76.76 & 70.84 \\ \bottomrule
\end{tabular}
}
\caption{Main results on GrailQA.}
\label{tab:main_results_grailqa}
\end{table}

\subsection{Dataset \& Preprocessing}

\textbf{WebQuestionsSP} (WebQSP) \cite{Yih-Wen-tau-ACL-2016-WebQSP}, \textbf{ComplexWebQuestions 1.1} (CWQ) \cite{Talmor-Alon-NAACL-2018-ComplexWebQuestions-CWQ}, and \textbf{GrailQA} \cite{Gu-Yu-WWW-2021-GrailQA} are standard KBQA benchmarks on Freebase \cite{Bollacker-Kurt-SIGMOD-2008-Freebase}. They pair natural-language questions with SPARQL queries.
For WebQSP, questions can be categorized into 1-hop and 2-hop types according to the length of the inferential relation chain.
CWQ extends WebQSP by incorporating four types of complex questions: Conjunction (Conj), Composition (Compo), Comparative (Compa), and Superlative (Super).
For GrailQA, the dev split is organized into three generalization levels: i.i.d., compositional, and zero-shot.

\textbf{KQA Pro} \cite{Cao-Shulin-ACL-2022-KQAPro} is a large-scale dataset designed for complex question answering over a densely connected subset of Wikidata \cite{Vrandečić-Denny-CACM-2014-Wikidata}. It features nine types of complex questions, including Count (Ct), Query Attribute (QA), Query Attribute Qualifier (QAQ), Query Name (QN), Query Relation (QR), Query Relation Qualifier (QRQ), Select Among (SA), Select Between (SB), and Verify (Vf).

Statistics of the test sets are listed in Table~\ref{tab:data_statistics}. We evaluate WebQSP on the full preprocessed test set. For CWQ, GrailQA, and KQA Pro, we randomly sample 500, 500, and 900 questions, respectively, before inference. These fixed random samples preserve the original test-set distribution in expectation and are shared by all methods. The samples are not filtered by model output, difficulty, answerability, or question length.

\subsection{Baselines}

For the \textbf{low-resource setting}, we compare against linear decision-making and search baselines. Interactive-KBQA \cite{Xiong-Guanming-ACL-2024-Interactive-KBQA} is the linear baseline, and we also report its five-run majority vote. For tree traversal, we use BFS, DFS, and Random Walk, which treat all nodes equally. IG-Beam Search uses the same actor, tools, sanitization, and IG reward as Fast MCTS, but replaces UCT with level-wise beam pruning. Fast MCTS \emph{w/ Const Reward} is an ablation that keeps the Fast MCTS search procedure but replaces the IG score with a constant reward for every child, isolating the contribution of IG scoring.

We include KBQA-o1 \cite{Luo-Haoran-arXiv-2025-KBQA-o1} as the closest prior MCTS-style KBQA method.\footnote{Its GrailQA setting uses 40 annotated trajectories plus large-scale unlabeled exploration and incremental fine-tuning, so we do not treat it as directly comparable.}

For \textbf{reward design}, we compare CoT prompting \cite{Wei-Jason-NeurIPS-2022-ChainOfThought} with a logical-form reward model fine-tuned to generate SPARQL, following KBQA-o1. Details and broader scoring comparisons are in Appendices~\ref{app_sec:logic_form_reward_model} and~\ref{sec:comparison_of_different_scoring_methods}.

For \textbf{reference systems}, we include DeCAF \cite{Yu-Donghan-ICLR-2023-DecAF} and BART-SPARQL \cite{Cao-Shulin-ACL-2022-KQAPro}. DeCAF uses the authors' released inference results evaluated on the same subsets when applicable. These systems provide context but are not direct low-resource comparisons.

For \textbf{open-source LLMs}, we compare Llama-3.1-8B-Instruct \cite{Aaron-etal-arXiv-2024-Llama3}, Qwen2.5-7B-Instruct \cite{An-etal-arXiv-2024-Qwen2.5}, and gemma-2-9b-it \cite{Gemma-Team-arXiv-2024-Gemma2}. By default, the SFT fine-tuned Llama-3.1-8B-Instruct is the actor and its original instruction-tuned version is the reward model.

\subsection{Implementation Details}
\label{sec:imp_details}

For comparability with prior work, we follow the Interactive-KBQA protocol on WebQSP, CWQ, and KQA Pro, and additionally include GrailQA.
We fine-tune the actor on manually annotated trajectories with intermediate reasoning steps. The training objective and data collection process are given in Appendix~\ref{app_sec:llm_fine_tuning}. The GrailQA annotation protocol is documented in Appendix~\ref{app_sec:grailqa_annotation}.

Unless otherwise stated, all SFT and search-based methods use the SFT fine-tuned Llama-3.1-8B-Instruct as the actor. The GPT-4.1 prompting baseline (gpt-4.1-2025-04-14) is rerun by us under the same tool interface and evaluation subsets, using the same prompting protocol as the linear agent and no additional annotated trajectories beyond the low-resource demonstrations.

For MCTS search, we set the sampling width $n=5$, early-stopping threshold $k=5$, simulation budget $B=50$, maximum selection iterations $T=12$, decay coefficient $\gamma=0.1$, and expected depth $d_{exp}=7$. We use temperature 0.7 for actor sampling and set reward scale $\alpha=2$ and slope $\beta=10$ for all datasets.
The UCT exploration coefficient and all search hyperparameters are fixed across datasets without dataset-specific tuning; the sigmoid mapping keeps rewards bounded, making the fixed exploration scale stable in practice.

\subsection{Evaluation Metrics}

For semantic parsing-based methods that generate logical forms and produce unordered answer sets, we employ the F1 score as the primary evaluation metric. 
To provide comprehensive evaluation, we additionally report the Random Hits@1 (RHits@1) metric \cite{Shu-Yiheng-EMNLP-2022-TIARA} and the Exact Match (EM) score \cite{Talmor-Alon-NAACL-2018-ComplexWebQuestions-CWQ}. 
For the KQA Pro dataset, following \cite{Cao-Shulin-ACL-2022-KQAPro}, we report accuracy, which is defined as the exact one-to-one correspondence between the predicted and ground truth answer sets.

\subsection{Main Results}

\paragraph{Overall Performance.}
Tables~\ref{tab:main_results} and \ref{tab:main_results_grailqa} show that, with only several hundred annotated trajectories, Fast MCTS consistently outperforms linear baselines and generic tree traversal, confirming the value of explicit search under scarce supervision. It surpasses Classic MCTS on WebQSP, CWQ, and GrailQA, while remaining comparable on KQA Pro. The reference systems provide useful context, but should not be read as direct low-resource comparisons.

\paragraph{Comparison with Search Baselines.}
IG-Beam Search improves over some unguided traversal baselines but remains below Fast MCTS on all four benchmarks. This indicates that the gain does not come solely from using IG as a reranker. UCT-style adaptive exploration keeps uncertain branches explorable and revisits states according to accumulated evidence, while beam search makes irreversible top-$B$ pruning decisions at each depth and can discard branches whose value becomes clear only after later KB interactions. The gap between Fast MCTS and its ``w/ Const Reward'' ablation (e.g., 70.15 vs.\ 57.99 on CWQ) further shows that informative intermediate scoring is essential; the Fast MCTS search skeleton alone is not sufficient.

\subsection{Comparison with KBQA-o1}

\begin{table}[ht]
\centering
\scalebox{1}{
\begin{tabular}{lcc}
    \toprule
    \textbf{Model} & \textbf{KBQA-o1} & \textbf{Fast MCTS} \\ \midrule
    Llama-3.1-8B & 59.8 & 66.11 \\
    Qwen2.5-7B & 57.8 & 66.19 \\
    Gemma-2-9B & 58.9 & 63.22 \\ \bottomrule
\end{tabular}
}
\caption{Comparison with KBQA-o1 on the full WebQSP test set under the 100-shot setting. KBQA-o1 results are taken from the original paper; Fast MCTS is run by us under the same 100-shot setting.}
\label{tab:kbqa_o1_comparison}
\end{table}

KBQA-o1 is the closest prior method under the 100-shot WebQSP setting because it also combines tree search with learned value estimation. The key difference is where supervision enters search: KBQA-o1 trains a reward model from terminal outcomes, whereas Fast MCTS uses an IG reward as a dense process-level signal at each expansion step. Under the same 100-shot setting and full WebQSP test set, Table~\ref{tab:kbqa_o1_comparison} shows that Fast MCTS consistently outperforms KBQA-o1 across all three backbone models. We include the reported GrailQA result of KBQA-o1 in Table~\ref{tab:main_results_grailqa} as a reference, but treat the aligned direct comparison as limited to WebQSP because the other KBQA-o1 settings use different data construction and training protocols.

\subsection{Accuracy-Cost Trade-off}

\begin{table}[ht]
\centering
\scalebox{1}{
\begin{tabular}{lcccc}
    \toprule
    \textbf{Method} & \textbf{F1} & \textbf{\#Actor} & \textbf{Time} & \textbf{RTime} \\ \midrule
    \multicolumn{5}{c}{CWQ} \\ \midrule
    Majority Vote & 55.97 & 32.44 & 140.67 & 1.00x \\
    Classic MCTS & 63.44 & 55.15 & 243.43 & 1.73x \\
    Fast MCTS & 70.15 & 27.49 & 203.84 & 1.45x \\ \midrule
    \textbf{Method} & \textbf{Acc} & \textbf{\#Actor} & \textbf{Time} & \textbf{RTime} \\ \midrule
    \multicolumn{5}{c}{KQA Pro} \\ \midrule
    Majority Vote & 83.44 & 30.49 & 98.55 & 1.00x \\
    Classic MCTS & 86.67 & 44.13 & 154.85 & 1.57x \\
    Fast MCTS & 86.44 & 20.77 & 121.77 & 1.24x \\ \bottomrule
\end{tabular}
}
\caption{Computational efficiency comparison. \#Actor counts autoregressive actor-generation calls only. Time is average inference time per question in seconds, including reward forward passes and tool execution. Majority Vote is obtained from five independent runs. RTime: relative time vs.\ Majority Vote.}
\label{tab:computation_comparison}
\end{table}

Table~\ref{tab:computation_comparison} reports representative efficiency results on CWQ and KQA Pro, covering complex Freebase and heterogeneous Wikidata settings. Fast MCTS uses fewer actor calls and lower wall-clock time than rollout-based Classic MCTS on both datasets, while achieving higher accuracy on CWQ and comparable accuracy on KQA Pro. It is still slower than Majority Vote in wall-clock time because each expansion requires forward-pass reward evaluation and the search explores a broader state space. Thus, our efficiency claim is a better accuracy-cost trade-off relative to Classic MCTS, not absolute wall-clock dominance over simple linear inference.

\subsection{Validation of IG as Value Proxy}
\label{sec:ig_validation}

\begin{table}[ht]
\centering
\scalebox{1}{
\begin{tabular}{lcccc}
    \toprule
    \textbf{Method} & \textbf{R@1} & \textbf{R@2} & \textbf{R@3} & \textbf{Kendall's $\tau_{b}$} \\ \midrule
    \multicolumn{5}{c}{CWQ} \\ \midrule
    IG  & 75.73 & 99.03 & 100 & 0.37 \\
    CoT & 62.14 & 92.23 & 98.06  & 0.12 \\ \midrule
    \multicolumn{5}{c}{KQA Pro} \\ \midrule
    IG  & 68.46 & 95.97 & 100 & 0.11 \\
    CoT & 66.44 & 85.91 & 93.29  & 0.19 \\ \bottomrule
\end{tabular}
}
\caption{Consistency between IG/CoT and true reward within parent nodes. R@k denotes Recall@k (\%). Kendall's $\tau_b$ is reported on the $[-1,1]$ scale.}
\label{tab:ig_cot_consistency}
\end{table}

Fast MCTS assumes that the immediate IG score preserves promising sibling branches well enough to replace rollout-based value estimation. We therefore compare proxy rankings against Monte-Carlo value estimates obtained from terminal rewards, which serve as a value oracle for sibling-level ranking.

\paragraph{Experimental Design.}
For each question, we run Classic MCTS, execute terminal SPARQL queries, compute F1 against gold answers, and backpropagate these terminal F1 values to obtain per-node Monte-Carlo value estimates (referred to as \textbf{true rewards} below for brevity). We then compare IG and CoT rankings within each sibling set. We use \textbf{Recall@k} as the primary metric, measuring whether the top-$k$ proxy ranking retains the best true-value child, and report \textbf{Kendall's $\tau_b$} \cite{Kendall-Biometrika-1945-Treatment-of-Ties} as a stricter full-ranking diagnostic.

\paragraph{Results and Analysis.}
Table~\ref{tab:ig_cot_consistency} shows that IG consistently outperforms CoT in Recall@k, indicating better preservation of high-value children. Kendall's $\tau_b$ is mixed: IG is higher on CWQ, while CoT is higher on heterogeneous KQA Pro. This supports our intended use case: IG is not a calibrated total-order estimator, but a branch-preserving proxy. Its 100\% Recall@3 on both datasets explains why Fast MCTS remains close to Classic MCTS on KQA Pro while being much cheaper.

\subsection{Robustness to Search Stochasticity}

\begin{table}[ht]
\centering
\scalebox{1}{
\begin{tabular}{llcc}
    \toprule
    \textbf{Dataset} & \textbf{Method} & \textbf{Mean $\pm$ Std} & \textbf{95\% CI} \\ \midrule
    CWQ & Classic MCTS & 62.12 $\pm$ 1.95 & [59.70, 64.54] \\
    CWQ & Fast MCTS & 69.46 $\pm$ 1.14 & [68.04, 70.88] \\
    KQA Pro & Classic MCTS & 87.89 $\pm$ 1.89 & [85.54, 90.24] \\
    KQA Pro & Fast MCTS & 87.52 $\pm$ 1.62 & [85.51, 89.53] \\ \bottomrule
\end{tabular}
}
\caption{Multi-seed robustness of Classic MCTS and Fast MCTS on additionally sampled 100-question subsets. We report F1 on CWQ and accuracy on KQA Pro.}
\label{tab:multi_seed_robustness}
\end{table}

Because search and LLM sampling are stochastic, we rerun Classic MCTS and Fast MCTS with five random seeds on additionally sampled CWQ and KQA Pro subsets. Table~\ref{tab:multi_seed_robustness} shows that Fast MCTS remains clearly better on CWQ and comparable to Classic MCTS on KQA Pro. This supports our main conclusion while preserving the more precise claim that Fast MCTS improves the accuracy-cost trade-off rather than universally dominating Classic MCTS in raw accuracy.

\subsection{When Does Fast MCTS Help?}

\begin{figure}[htbp]
    \centering
    \includegraphics[width=0.49\columnwidth]{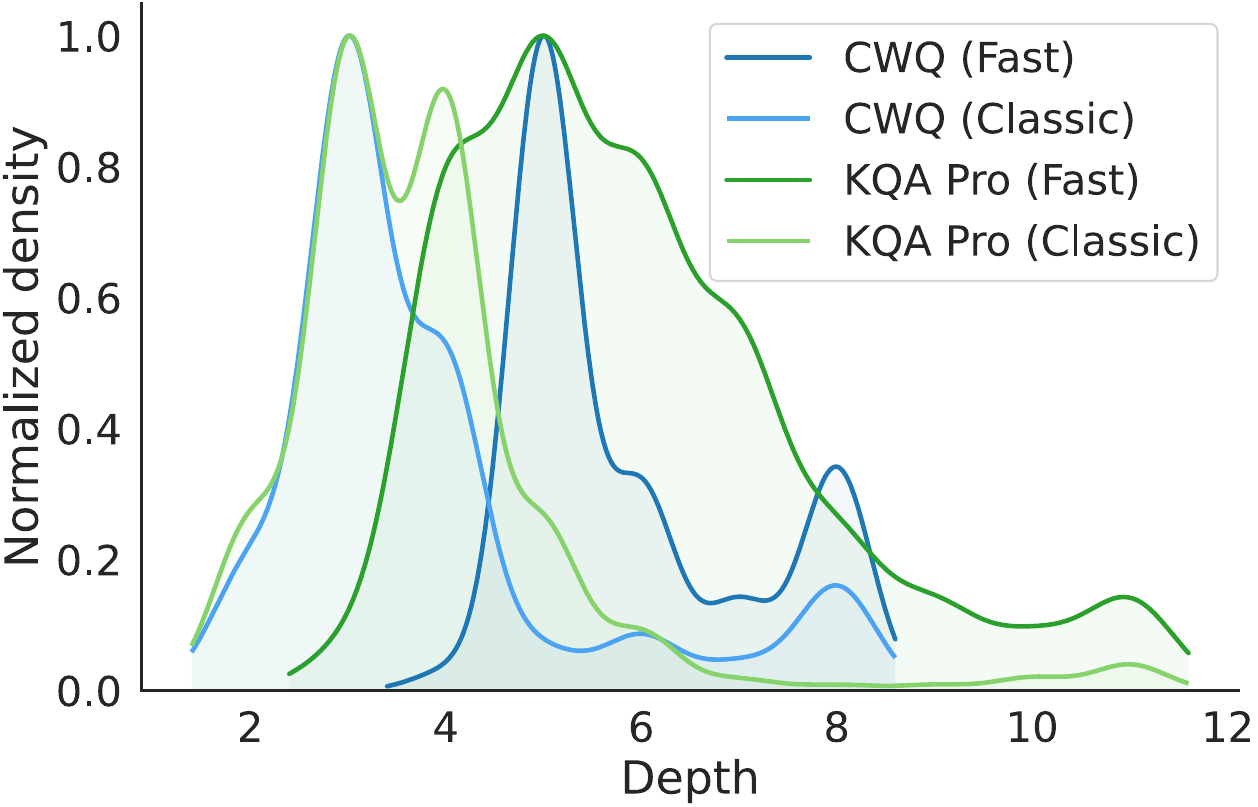}
    \includegraphics[width=0.49\columnwidth]{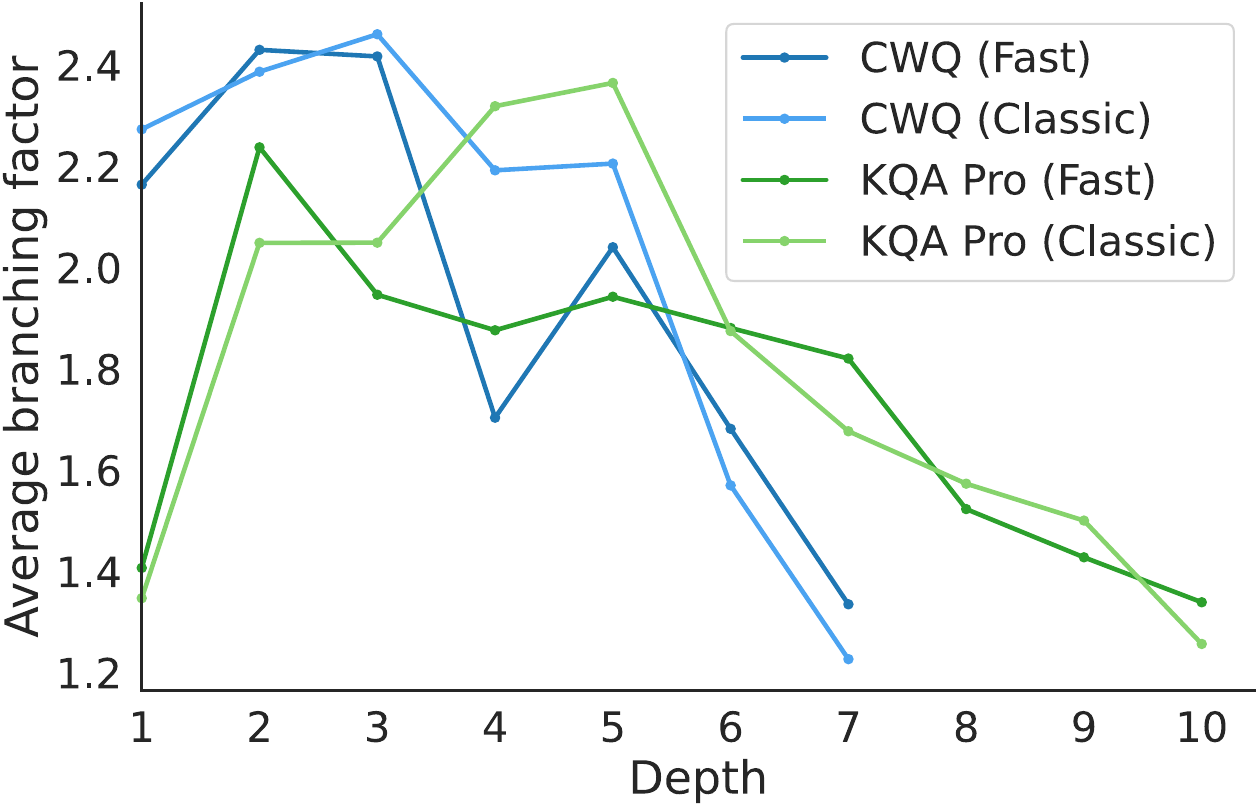}
    \caption{Search-tree depth distribution (left) and average branching factor by depth (right).}
    \label{fig:tree_statistics}
\end{figure}

Figure~\ref{fig:tree_statistics} shows substantial search-space differences across datasets. On CWQ, the space is narrower and dominated by predicate matching over crowd-sourced Freebase relations, where many candidates are lexically close but few are execution-consistent; here IG is effective because wrong branches lose score quickly after exposing mismatched predicates. KQA Pro is different: children may represent qualifiers, comparisons, counts, boolean operators, or deep filters, yielding a wider and more heterogeneous frontier. Local progress still helps pruning, but is a weaker proxy for final executability, so rollout-based Classic MCTS remains on par with or slightly above IG-based scoring. Overall, Fast MCTS is most useful when local tool observations strongly predict downstream success, while Classic MCTS remains valuable when long-horizon heterogeneous operators dominate.

\subsection{Error Analysis}

\begin{table}[htbp]
\centering
\scalebox{1}{
    \begin{tabular}{lccc}
        \toprule
        \multicolumn{1}{c}{\textbf{Metric/Type}} & \textbf{CWQ} & \textbf{GrailQA} & \textbf{KQA Pro} \\ \midrule
        MaxF1@5 & 76.29 & 85.46 & 91.40 \\
        Empty\%@5 & 16.69 & 14.21 & 8.13 \\ \midrule
        \multicolumn{4}{c}{Mismatch} \\ \midrule
        Golden Wrong & 5 & 10 & 4 \\
        Ambiguous & 34 & 2 & 12 \\ \midrule
        \multicolumn{4}{c}{Error} \\ \midrule
        Entity Linking & 9 & 26 & 6 \\
        Predicate Search & 21 & 4 & 0 \\
        Reasoning & 25 & 58 & 52 \\
        Other & 6 & 0 & 26 \\ \bottomrule
    \end{tabular}
}
\caption{Error analysis. MaxF1@5 is the best F1 among the top 5 predictions; Empty\%@5 is the share of cases where all top 5 predictions score zero.}
\label{tab:error_analysis}
\end{table}

We use MaxF1@5, the best F1 among the top 5 predictions, to estimate the search upper bound, and Empty\%@5 to measure the share of cases where all 5 predictions score zero. We manually analyze a random sample of 100 errors from each dataset, as shown in Table~\ref{tab:error_analysis}. We divide cases into \textbf{Mismatch} (semantically correct but different from the gold SPARQL) and \textbf{Error} (incorrect generation).

Across datasets, the gap between MaxF1@5 and final performance indicates that search often reaches promising branches but fails to rank or complete them into executable, type-consistent queries. CWQ has the highest Ambiguous rate, suggesting stronger evaluation noise from multiple valid Freebase reformulations, whereas GrailQA and KQA Pro more often reflect genuine modeling errors; Golden Wrong cases remain in all three datasets. Among true generation errors, CWQ is characterized by a relatively high share of predicate-retrieval errors alongside reasoning errors, GrailQA is dominated by reasoning and entity linking, and KQA Pro by reasoning and qualified-structure generation.

Concretely, \textbf{Golden Wrong} denotes cases where the gold SPARQL is itself inconsistent with the question. \textbf{Ambiguous} denotes alternative correct logical forms under single-reference evaluation, e.g., for ``what country did buddha come from?'', the gold uses ``nationality'' while our predicted ``place\_of\_birth'' is also correct. \textbf{Entity Linking} denotes choosing the wrong entity, e.g., ``Martin Luther King'' instead of ``Martin Luther King, Jr.''. \textbf{Predicate Search} denotes failure to retrieve the required schema edge, e.g., missing ``people.person.employment\_history $\rightarrow$ business.employment\_tenure. company'' for ``where did rudolf virchow conduct his research''. \textbf{Reasoning} denotes errors in operator choice, composition, or predicate direction. \textbf{Other} covers remaining failures such as formatting errors, especially for qualified Wikidata structures. These patterns suggest stronger predicate retrieval for CWQ, more robust disambiguation for GrailQA, and more structure-aware reasoning for KQA Pro.

\section{Conclusion}
We presented an MCTS framework for KBQA with an IG reward computed from forward passes of an open-source LLM. Across four benchmarks, the method substantially improves over linear baselines in low-resource settings. Fast MCTS uses the reward as a process-level branch-ranking proxy and generally provides a better accuracy-cost trade-off than rollout-based Classic MCTS, with the main accuracy gap appearing on the more heterogeneous KQA Pro benchmark.

\section*{Limitations}
Our study has several limitations. Fast MCTS is more efficient than Classic MCTS but not always faster than majority-vote linear inference because reward evaluation adds forward passes. The IG reward ranks sibling nodes rather than estimating calibrated terminal values, so it depends on whether local observations predict downstream success. Finally, except on the full WebQSP test set, we use sampled test sets and dataset-specific actors, so we do not claim cross-KB transfer or large-scale production robustness.

\begin{acks}
This work was supported by the National Key Research and Development Program of China (No. 2024YFC3307301).
\end{acks}

\appendix
\section{Appendix}
This appendix provides additional details for the reward design, scoring baselines, and implementation.

\subsection{Details of Information Gain Reward}
\label{app_sec:ig_reward_details}

\paragraph{Perplexity Definition.}
For a token sequence $X=(x_1,\ldots,x_T)$ and a language model with parameters $\theta$, we compute
\begin{equation}
    \operatorname{PPL}(X) = \exp\Big(- \frac{1}{T} \sum_{t=1}^{T} \log p_{\theta}(x_t \mid x_{<t})\Big),
\end{equation}
where lower perplexity indicates higher model-assigned probability. Throughout this paper, $\log$ denotes the natural logarithm. The same frozen base model is used for $\operatorname{PPL}(H')$ and $\operatorname{PPL}(H',Q)$; no additional fine-tuning is performed for the reward model.

\paragraph{Mutual-information-inspired motivation.}
A natural motivation for our guidance signal is the mutual information between the question and the sanitized history:
\begin{equation}
    \begin{aligned}
        I(Q;H') &= \log p(Q|H') - \log p(Q) \\
                &= \log p(H',Q) - \log p(H') - \log p(Q).
    \end{aligned}
\end{equation}
Since $\log p(Q)$ is constant across nodes, ordering depends only on $\log p(H',Q) - \log p(H')$. Using the language model as a density estimator, we approximate
\begin{equation}
    \log p(X) \approx -|X|\,\log \operatorname{PPL}(X),
\end{equation}
where $|X|$ denotes the token length of sequence $X$. Let $(H',Q)$ denote the token-level concatenation of $H'$ and $Q$, so that $|(H',Q)| = |H'| + |Q|$. Substituting gives
\begin{equation}
    \begin{aligned}
        I(Q;H') &\approx -\bigl(|H'|+|Q|\bigr)\,\log \operatorname{PPL}(H',Q) \\
                &\quad + |H'|\,\log \operatorname{PPL}(H').
    \end{aligned}
\end{equation}
This motivates the following unnormalized, length-aware proxy:

\begin{equation}
    \begin{aligned}
        \mathrm{IG}_{\text{proxy}}(H', Q) &= -\bigl(|H'|+|Q|\bigr)\,\log \operatorname{PPL}(H',Q) \\
                                       &\quad + |H'|\,\log \operatorname{PPL}(H').
    \end{aligned}
    \label{eq:ig_proxy}
\end{equation}
In the final reward, we use the simpler ratio form
\begin{equation}
    R(H', Q) = \frac{\operatorname{PPL}(H')}{\operatorname{PPL}(H',Q)}.
\end{equation}
This ratio differs from $\mathrm{IG}_{\text{proxy}}(H', Q)$ in that the length factors are dropped. We therefore treat it as a practical question-conditioned PPL surrogate, not as a strict mutual-information estimator. The score is used only for local branch ranking and is empirically validated in Section~\ref{sec:ig_validation}, rather than being assumed to provide calibrated information-gain values.

\paragraph{Sanitization Details.}
Sanitization removes lexical redundancy that could artificially inflate the PPL-ratio score while preserving structural reasoning progress. Specifically, we (i) eliminate all free-form \emph{thought} texts, which often paraphrase the question; (ii) for \texttt{SearchNodes} actions, retain only the action type while removing lexical parameters such as copied entity mentions; (iii) preserve structured parameters for later actions, since they are derived from tool observations rather than directly copied from the question; and (iv) retain only the final observation $o_d$, which is the outcome of the current action and provides the local evidence needed to assess that step.

The three tools in our agent, namely entity localization, predicate acquisition, and SPARQL construction, capture complementary aspects of question semantics. As a result, even an incomplete history can already contain informative evidence about whether the search is moving toward the correct logical form. The IG reward is therefore used to rank sibling states well enough for UCT to retain promising branches, not to exactly recover terminal rewards at every step.

\subsection{Details of Logic Form Reward Model}
\label{app_sec:logic_form_reward_model}

The reward model aims to assess the entire trajectory by scoring the final logical form. We use the question $Q$ as input, and the logical form (SPARQL) $S$ from the annotated training set $\mathcal{D}$ as output, forming SFT data for training the reward model $\pi$:
\begin{equation}
\mathcal{L}_{\text{SFT}}(\pi, \mathcal{D}) = -\mathbb{E}_{\mathcal{D}}[\log \pi(S | Q)].
\end{equation}

Moreover, the scoring function $R_\pi$ is defined as the sequence log-likelihood assigned by the LLM $\pi$ to a predicted logical form $S'$ given the question $Q$:
\begin{equation}
R_\pi(S' | Q) = \log \pi(S' | Q).
\end{equation}

\subsection{Comparison of Different Scoring Methods}
\label{sec:comparison_of_different_scoring_methods}

\begin{table}[ht]
\centering
\scalebox{0.9}{
\begin{tabular}{lcccc}
    \toprule
    \textbf{Method} & \textbf{WebQSP} & \textbf{CWQ} & \textbf{KQA Pro} & \textbf{Time} \\ \midrule
    CoT & 73.23 & 69.32 & 87.56 & 15.76x \\
    Logic Form & 73.09 & 69.23 & 86.91 & 0.61x \\
    IG & \multicolumn{1}{l}{} & \multicolumn{1}{l}{} & \multicolumn{1}{l}{} & \multicolumn{1}{l}{} \\
    \quad w/ Llama-3.1-8B & 69.14 & 63.44 & 86.67 & \multirow{3}{*}{1.00x} \\
    \quad w/ Qwen2.5-7B & 72.02 & 68.16 & 86.30 &  \\
    \quad w/ Gemma-2-9B & 71.73 & 68.52 & 85.75 & \\ \bottomrule
\end{tabular}
}
\caption{Comparison of different scoring methods in Classic MCTS. All experiments use instruction-tuned models, with model versions omitted for brevity.}
\label{tab:reward_method_analysis}
\end{table}

Table~\ref{tab:reward_method_analysis} compares our proposed IG reward with other methods. As shown in Table~\ref{tab:main_results}, even a Random Walk baseline is competitive with BFS/DFS and often improves over linear inference, suggesting that a reward need not perfectly estimate terminal return to be useful—which also explains why zero-shot CoT can provide a workable signal.
However, CoT is generative and roughly 15$\times$ slower than forward-pass-based approaches, and Classic MCTS calls the reward function $\sim$10 times per question, making such latency impractical.

The logical form reward is the most efficient, requiring only the question and final SPARQL as input, but it applies only after a complete logical form is available and is thus unsuitable as a dense reward for intermediate Fast MCTS states. It also suffers from surface-form mismatch: a predicted SPARQL that is semantically correct may still score low if it differs lexically from the gold annotation—e.g., ``\texttt{SELECT ?x WHERE \{"Honolulu" office\_holder ?x.\}}'' for ``who is the mayor of Obama's birthplace?''

In contrast, our IG reward leverages the full reasoning history, and the comparison across open-source reward models shows that this value proxy transfers reasonably well.

\subsection{Open-source LLM Fine-tuning \& Deployment}
\label{app_sec:llm_fine_tuning}

\begin{table}[ht]
\centering
\begin{tabular}{lc}
  \toprule
  \multicolumn{1}{c}{\textbf{Parameter}} & \textbf{Value} \\ \midrule
  Batch Size (per GPU) & 8 (2) \\
  Gradient Accumulation Step & 2 \\
  Model Max Length & 4,096 \\
  Learning Rate & 1e-5 \\
  Weight Decay & 1e-3 \\
  Epoch & 5 \\
  Warm Up Step & 0 \\
  Zero Stage & 3 \\ \bottomrule
\end{tabular}
\caption{Hyperparameter settings for fine-tuning.}
\label{app_tab:hyper_deepspeed}
\end{table}

We use Llama-3.1-8B-Instruct \cite{Aaron-etal-arXiv-2024-Llama3} as the base model. The LLM agent is supervised fine-tuned with the hyperparameters in Table~\ref{app_tab:hyper_deepspeed}; the reward model uses the original instruction-tuned weights without additional training.

The actor is fine-tuned on annotated trajectories with intermediate reasoning steps. At each step, the question and prior observations serve as context (masked in the loss); we optimize only the generated \textit{thought} and \textit{action} segments with the standard NLL objective.

To preserve comparability with Interactive-KBQA, we keep its Freebase fine-tuning setup unchanged and train one agent on merged WebQSP/CWQ annotations (300 trajectories). GrailQA is trained separately because its additional \texttt{SearchTypes} tool changes the action space; we newly annotate 150 GrailQA trajectories and fine-tune a dedicated agent with the same base model. KQA Pro also uses a separate agent (450 annotated trajectories) due to its Wikidata backend.

Training uses DeepSpeed~\cite{Rasley-Jeff-KDD-2020-DeepSpeed} and inference uses vLLM~\cite{Kwon-Woosuk-SOSP-2023-vLLM}, all on 4 NVIDIA A100 80G GPUs.

\subsection{Annotation Protocol for GrailQA Trajectories}
\label{app_sec:grailqa_annotation}

The GrailQA trajectory annotations used in this work were created by the authors through prompting a frontier closed-source LLM via its public API, followed by manual verification of each trajectory on a case-by-case basis to ensure correctness. This procedure was adopted to construct intermediate reasoning traces while maintaining strict faithfulness to the observed tool interactions.

Our manual verification followed two criteria. First, each step must rely only on information that had already been obtained from the historical observations, without introducing unsupported facts or fabricated constraints. Second, the final answer SPARQL must be semantically consistent with the original question, including its intended entities, relations, and constraints.

\section*{GenAI Usage Disclosure}
We used generative AI tools to polish the writing of this paper and to assist with annotating trajectories for the GrailQA dataset. All technical content, experimental design, results, and conclusions were produced and verified by the authors.

\bibliographystyle{ACM-Reference-Format}
\bibliography{mcts_kbqa}

\end{document}